\documentclass[10pt,twocolumn,letterpaper]{article}

\usepackage{iccv}
\usepackage{times}
\usepackage{epsfig}
\usepackage{graphicx}
\usepackage{amsmath}
\usepackage{amssymb}
\usepackage{subcaption}
\usepackage[autolanguage]{numprint}

\iccvfinalcopy
\def\iccvPaperID{7807} 

\ificcvfinal\fi

\newcommand{\logg}[1]{\log\!\left(\strut #1\right)}
\newcommand{\expect}[2]{\mathbf{E}_{#1\!}\left[ \strut #2 \right]}
\newcommand{\kl}[2]{\mathbf{KL}\!\left(\strut #1 \,\big\Vert\, #2\right)}
\DeclareMathAlphabet{\mat}{OT1}{cmss}{bx}{n}
\usepackage[pagebackref=true,breaklinks=true,colorlinks,bookmarks=false]{hyperref}
\begin{document}

%%%%%%%%% TITLE
\title{What Remains of Visual Semantic Embeddings}

\author{Yue Jiao\\
University of Southampton\\
Southampton, UK\\
{\tt\small yj5y15@soton.ac.uk}
\and
Jonathon Hare\\
University of Southampton\\
Southampton, UK\\
{\tt\small jsh2@ecs.soton.ac.uk}
\and
Adam Prügel-Bennett\\
University of Southampton\\
Southampton, UK\\
{\tt\small apb@ecs.soton.ac.uk}
}
\maketitle
%\thispagestyle{empty}

%%%%%%%%% ABSTRACT
\begin{abstract}
Zero shot learning (ZSL) has seen a surge in interest over the decade for its tight links with the mechanism making young children recognize novel objects. 
Although different paradigms of visual semantic embedding models are designed to align visual features and distributed word representations, it is unclear to what extent current ZSL models encode semantic information from distributed word representations.
In this work, we introduce the split of tieredImageNet to the ZSL task, in order to avoid the structural flaws in the standard ImageNet benchmark.  
We build a unified framework for ZSL with contrastive learning as pre-training, which guarantees no semantic information leakage and encourages linearly separable visual features.   
Our work makes it fair for evaluating visual semantic embedding models on
a ZSL setting in which semantic inference is decisive.
With this framework, we show that current ZSL models struggle with encoding semantic relationships from word analogy and word hierarchy. 
Our analyses provide motivation for exploring the role of context language representations in ZSL tasks.

\end{abstract}

\section{Introduction}

Zero-Shot Learning (ZSL), which aims to recognize unseen classes, is considered one of the most difficult generalization tasks.
ZSL is also an important task for demonstrating how a machine learning model understands high-level semantic information and transfers knowledge from seen to unseen classes.

After a decade of research, ZSL models have been shown to have made huge progress on some small and medium sized datasets annotated with handmade attributes~\cite{xian2017zero, schonfeld2019generalized, han2020learning}.
The key to transferring semantic knowledge from seen to unseen classes is often achieved by building a visual semantic embedding model, which aims to build a bridge to fill the gap between visual information and semantic information~\cite{lampert2009learning, socher2013zero, frome2013devise}.
This approach is based on the hypothesis that similar structural relationships emerge from independent visual and linguistic representations~\cite{roads2020learning,ilharco2020probing}.

ZSL methods  fail dramatically when they are tested on the ImageNet dataset~\cite{imagenet_cvpr09}. To address this problem, recent works propose to introduce different mechanisms to represent robust semantic hierarchy~\cite{wang2018zero, kampffmeyer2019rethinking, liu2020hyperbolic}.
However, there is still an open question: \emph{do current ZSL models perform useful semantic inference on the ImageNet dataset?} 
Exploring the role of distributed semantics and visual semantic alignment models in the ZSL task is still lacking. 
For example, if we have two images: one of a working dog and one of a hunting dog, it is not clear whether the semantic difference and the semantic hierarchy between ``working'' and ``hunting'' will help us distinguish these two images.    

\begin{figure}[tbp]
\begin{center}
   \includegraphics[width=\columnwidth]{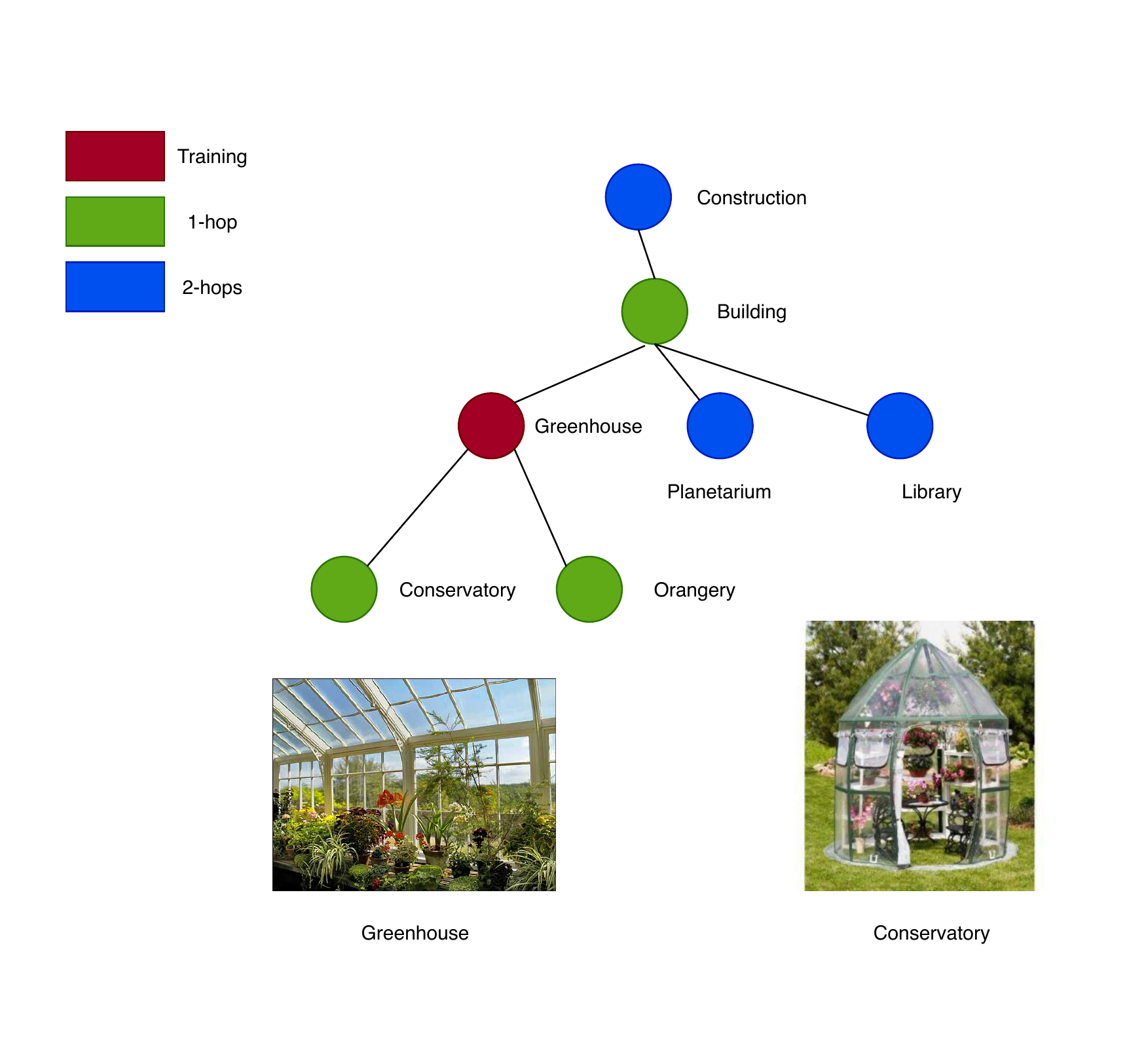}
\end{center}
   \caption{Illustration the standard ZSL ImageNet data split over the Wordnet hierarchy. The red node ``greenhouse'' is in the seen classes, while its parent ``building'' and children ``conservatory'' and ``orangery'' are tested as the unseen classes.
   The standard ZSL aims to distinguish objects from their hyponyms and hyponyms making it difficult to demonstrate whether ZSL models can learn semantic concepts via semantic embedding and alignment.}
\label{fig:stand}
\end{figure}

As Hascoet \etal~\cite{hascoet2019zero} highlight, the standard ImageNet benchmark proposed by Frome \etal~\cite{frome2013devise} has structural flaws due to its data splitting:  although the full set of classes is split into disjoint training and test sets, the test classes are hypernyms or hyponyms of the training classes within the Wordnet~\cite{miller1995wordnet} hierarchy. 
When an image of a ``greenhouse'' is classified as a ``building'' or ``conservatory'' (see Figure~\ref{fig:stand}), in the traditional ZSL setting, we consider it is a case of classification error, whilst in terms of the semantic definition of a ``greenhouse'', it doesn't seem to be an error.
On the other hand, in the standard benchmark, test classes, which are out of the 1000 classes of the ILSVRC 2012 1K~\cite{russakovsky2015imagenet},
are imbalanced. Some categories have the high sample populations, while some  categories only have low samples. 
Therefore, the standard benchmark is not fair enough to demonstrate to what extent visual semantic embedding models are affecting semantic understanding in the ZSL task.
To address this problem, we propose using the split of tieredImageNet~\cite{ren2018meta}, a large subset of ILSVRC 2012 1K, to build a new ZSL benchmark. 
In the tieredImageNet dataset, classes are grouped into categories corresponding to higher-level nodes in the ImageNet hierarchy.
These categories are split into two parts to ensure that there is no hyponymy and hypernymy between the training and test classes, thus ensuring that the two sets are semantically and linguistically disjoint.
There will be no longer any super classes in the testing set.
The tieredImageNet split makes ZSL a fair game in which models
depend on visual-semantic alignment.

With the new tieredImageNet split, we test four main paradigms to learning a visual semantic embedding model.
The first paradigm is to map pre-trained image features into a rich word embedding space, such as Word2Vec~\cite{mikolov2013distributed} or GloVe~\cite{pennington2014GloVe}. The typical work in this paradigm is done by Frome \etal~\cite{frome2013devise}, which helps visual object categorization systems handle very large numbers of labels.    
The second paradigm is to utilize an Autoencoder (AE) or a Variational Autoencoder (VAE) to explore a latent semantic space of image features, in to which word vectors are jointly mapped. Semantic AE~\cite{sung2018learning}, Cross and Distribution Aligned VAE~\cite{schonfeld2019generalized} and SCAN~\cite{higgins2017scan} belong to this paradigm.
The third paradigm uses explicit knowledge graphs which encode relationships between object classes. Like~\cite{wang2018zero}, the weights and biases of the final linear layer of the pre-trained image classifier are the new learning objectives. 
The output node embeddings from a graph convolutional neural network (GCN) are used to predict the visual classifier for each category, when corresponding word vectors are inputted. 
The forth paradigm considers the visual semantic embedding as a hyperbolic space and maps visual features and word vectors to a Poincar\'e disk.
This kind of ZSL methods, proposed by Liu \etal~\cite{liu2020hyperbolic}, 
follows the empirical rule that a hyperbolic latent space can yield more interpretable representations if the data has hierarchical structure~\cite{chamberlain2017neural, mathieu2019continuous}.  Compared with the 
first paradigm, the fourth employs an exponential map instead of a nonlinear 
map in the Euclidean latent space.

To avoid supervised pre-training of a state-of-the-art deep neural network for visual object recognition to create the visual feature extractor, we introduce Self-Supervised Learning (SSL)~\cite{jing2020self} into the framework of ZSL. SSL allows us to obtain general visual features without any explicit semantic information being introduced. We have many concerns with the pre-trained classifier towards our goal of understanding the role of visual semantic embeddings that contribute to recognize novel objects.
There are no guarantees that classifier trained using supervised methods on the seen classes will be able to generate separable image features for novel categories without any knowledge transfer like Deep Transfer Clustering~\cite{Han2019learning}.
However, a SSL encoder can provide separable image features without utilising any label information. 
Therefore, we believe that introducing SSL into the framework of ZSL can help
us understand how distributed semantics are aligned and how they affect ZSL models recognize novel classes.

Finally, within our proposed zero-shot setting, we come to the conclusion that current ZSL models struggle with semantic inference on the WordNet hierarchy.
When the unseen classes are not the hypernyms and hyponyms of seen classes, 
ZSL models with different semantic alignment mechanisms do not show the 
ability to use word analogy and word hierarchy implicitly, even though the image features of novel classes are near linearly separable.    
We show that graph-based visual semantic embedding models perform worse than the vanilla one which just learns a nonlinear map.

Therefore, we believe future ZSL frameworks 
should explore the role of contextual word representations on large vision-and-language tasks.

Our main contributions are to:
\begin{itemize}
  \item Introduce the tieredImagenet split into ZSL to replace the standard one, which avoid major structural flaws in ZSL benchmarking;
  \item Build a unified framework for ZSL with SSL pre-training, thus preventing any semantic information leakage. This makes the visual semantic 
  embedding models be the only decisive factor in the ZSL framework; 
  \item Demonstrate that current ZSL methods can not generalize from complicated semantic relationships. Encoded information from word analogy and word hierarchy is not enough for the ZSL task. 
\end{itemize}

\section{Preliminaries}
\subsection{Problem Definition}
Let $D_{S} = \{(x, y) \mid  x \in X_{S}, y \in  Y_{S} \}$ be a set of training examples consisting of seen images from $X_{S}$ and seen class labels from $Y_{S}$.
Let $D_{U} = \{(x, y) \mid  x \in X_{U}, y \in  Y_{U} \}$ be a set of testing examples. 
$Y_{U}$ represents the unseen set of class labels, which is disjoint from $Y_{S}$.
The task of ZSL is to learn a classifier, $X\rightarrow Y$, where $X$ is the union set of $X_{S}$ and $X_{U}$ and $Y$ is the union set of $Y_{S}$ and $Y_{U}$.

\subsection{Word Embeddings}
The idea of word embedding methods is derived from the distributional hypothesis in linguistics: words that are used and occur in the same contexts tend to purport similar meanings~\cite{harris1954distributional}.
Word representations learned in an unsupervised manner from the contextual relationship of words, or the co-occurrence statistics of words, in large text corpora are the main source of high-level semantic knowledge in most ZSL approaches.
In ZSL these word embedding models provide a mapping, $Y\rightarrow W$, from class labels to high dimensional word vectors. We denote the word vector corresponding to a label $y$ as $w(y)$.
Cosine similarity is usually used to predict word similarity. For a pair of word representations $w_{i}$ and $w_{j}$, we have:
$$\mathrm{similarity}(w_{i}, w_{j})=\frac{w_{i}\top w_{j} }{\left \| w_{i} \right \|\left \| w_{j} \right \|} \enspace .$$

\subsection{Contrastive Learning}
Previous ZSL models begin with training a classifier on the training set. This step leaks specific semantic information into the feature extractor component of a ZSL system. However, this step can not guarantee that we can obtain linear separable novel image features, therefore, it is very difficult to figure out 
whether the feature extractor or the downstream visual semantic embedding model affect the generalization performance. 
Contrastive learning (CL) provides a framework to learn representations of identity by pushing apart two views of different objects and bringing together two views of same objects in a representation space~\cite{chopra2005learning}.
Recent works show that CL can generate linear separable image features without any supervised label information~\cite{oord2018representation, tian2019contrastive, he2020momentum, chen2020simple}.
Given an anchor variables from the first view $v_{1,i}$, CL aims to score the correct positive variable from the second view $v_{2,i} \sim p(v_{2} \mid v_{1,i})$ higher compared to a set of $K$ negatives $v_{2,k} \sim p(v_{2})$.
A popular approach is to utilise the InfoNCE loss~\cite{oord2018representation}:
$$L_{NCE} = - \mathbf{E} \left [ \log \frac{\exp{h(f(v_{1,i}), f(v_{2,i}))}}{\sum _{j=1}^{K}\exp{h(f(v_{1,i}), f(v_{2,j}))}} \right ] $$
The function $h$ represents a critic head and the function $f$ is a shared encoder, which extracts view-invariant visual representations 
This way of feature learning is natural and powerful.
This allows the step of building the visual semantic embedding model to be the only
stage to align visual information and distributed linguistic representations.
When we freeze the visual encoder, we can evaluate what mechanism performs better on semantic understanding.

\subsection{Graph Convolutional Networks}
Graph convolutional networks (GCNs)~\cite{kipf2016semi} allow local graph operators to share the statistical strength between word vectors of classes, 
which is considered as a tool to utilize semantic information hidden in the WordNet hierarchy.

Given a graph with $N$ classes and a $d$ word embedding per class, $\mat{W}$ is the $N\times d$ word embedding matrix. We define a symmetric adjacency matrix $\mat{A}$ and a degree matrix $\mat{D}$ to represent the connections between the classes in the WordNet.
The propagation rule to perform convolutions on the graph is defined as:
\begin{align*}
    \mat{H}_{l+1} &= \sigma (\mat{D}^{-1} \mat{A} \mat{H}_{l} \mat{\Theta}_{l}),& \mat{H}_{0}&=\mat{W}
\end{align*}
Here $\mat{H}_{l}$ is the input at each layer of the GCN and $\mat{H}_{l+1}$ is the output. $\sigma$ is the activation function and $\mat{\Theta}_{l}$ is the learnable parameter matrix at the layer $l$.
We define a two layer GCN as $g$.

\subsection{Poincar\'e Embedding}
The Poincar{\'e} embedding model is another natural technique to capture hierarchical information in the WordNet hierarchy.
As it is not pre-trained on a large text corpus, each node in the Poincar{\'e} embedding model  does not contain any word co-occurrence statistics.
Poincar\'e embeddings preserve the distances between the nodes on the graph approximately, which can be used to compute semantic similarities.
Given two Poincar\'e embeddings $p_{i}$ and $p_{j}$, we have the distance:
$$ d(p_{i}, p_{j}) = \cosh^{-1}\left(1+2\frac{   \left \| p_{i}-p_{j} \right\| ^{2}}{  \left \| 1-p_{i} \right\| ^{2} + \left \| 1-p_{j} \right\| ^{2}    } \right)\;.$$

Two kinds of transformation are useful in a hyperbolic space.
The mapping $e$ to project a vector $v$ in Euclidean space to hyperbolic space is defined as:
$$ p = e(v) = \tanh(\left \| v \right\|) \frac{v}{\left \|v \right\|}$$
The Mobius multiplication $m$ is defined as:
$$m(x)= \tanh\left(\frac{\left \| mx \right \|}{\left \| x \right \|}\tanh^{-1}(\left \| x \right \|)\right)\frac{mx}{\left \| x \right \|}$$

\section{A Unified Framework of ZSL}
In this section, we provide our unified framework for the ZSL task. 
This new pipeline consists of two steps.

\begin{figure*}[t]
   \begin{subfigure}{.55\textwidth}
     \centering
     \includegraphics[height=7.3cm]{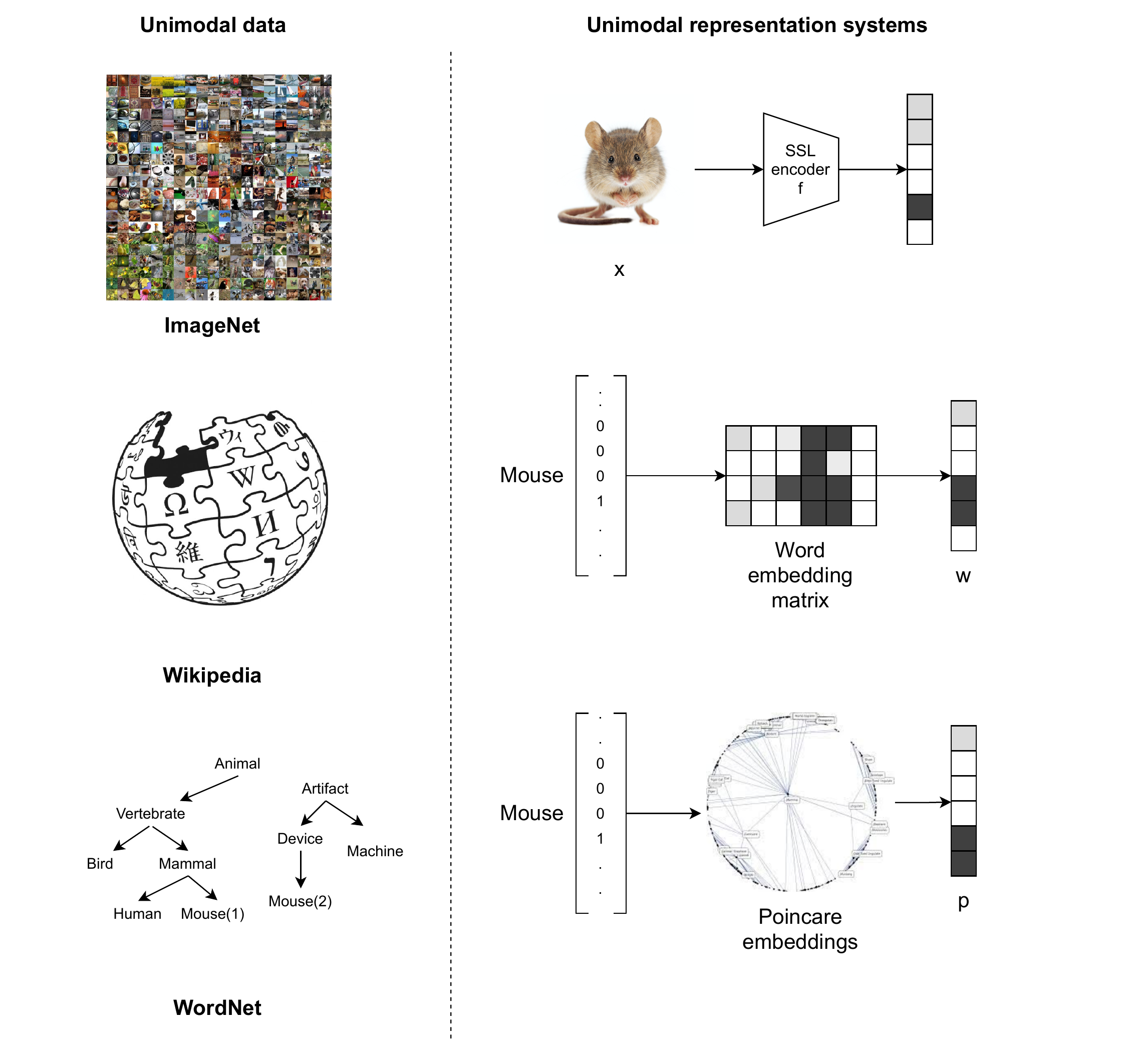}  
     \label{fig:one}
   \end{subfigure}\hfill
   \begin{subfigure}{.5\textwidth}
     \centering
     \includegraphics[height=7.5cm]{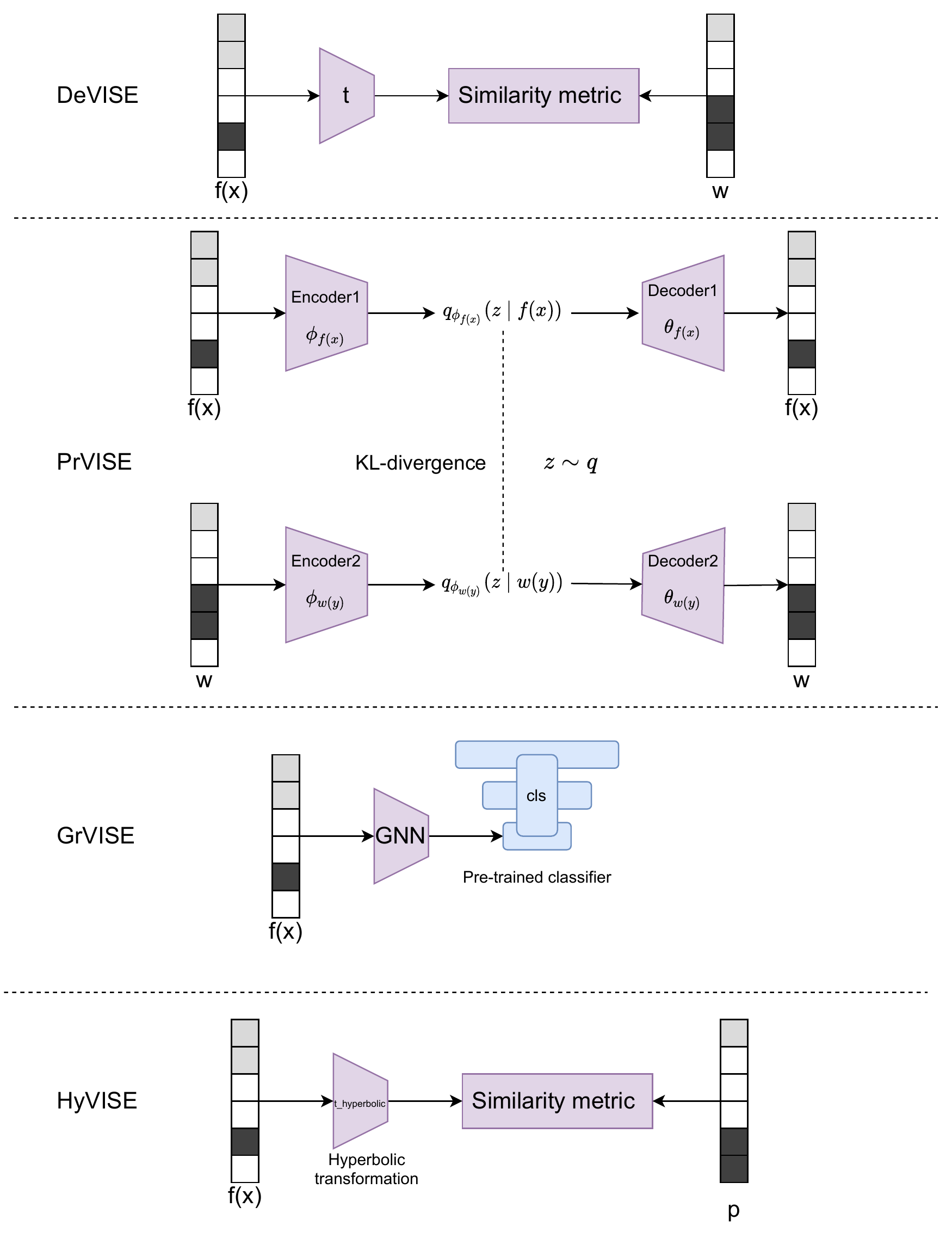}  
     \label{fig:two}
   \end{subfigure}
   \caption{Illustration of the process of building ZSL models in this work. There are two steps in the whole pipeline. Left: The first step is pre-training.
   In this step, an image encoder is learnt by Contrastive Learning on the ILSVRC 2012 1K. GloVe word embeddings and Poincar\'e embeddings are respectively trained on the Wikipedia corpus and the WordNet noun tree. (In this case, we use the subtree with the root node ``entity''.) Right: The second step is to learn a visual semantic embedding model, which can be seen as semantic alignment. In this framework we consider four independent paradigms including: DeVISE, PrVISE, GrVISE and HyVISE.  This two-step framework defines a game in which novel image features are linear
   separable in theory and visual semantic paradigms are the key to solve a semantic inference puzzle on the WordNet. Therefore, it is fairer to demonstrate how visual semantic embedding models do semantic generalization. }
   \label{fig:pipeline}
   \end{figure*}

\paragraph{Pre-training.} 
In this step, three unsupervised embedding systems are derived from the ILSVRC 2012 1K, the Wikipedia2014 \& Gigaword5 text corpus and the relation graph built from the WordNet.
The goal of build three embedding spaces is to capture  transferable visual, linguistic and structural information from different modalities,  which generalizes to an information alignment system.

After pre-training, image embeddings $I$ extracted by the CL encoder $f$, word embeddings $W$ and Poincar\'e embeddings $P$ are powerful features for the downstream visual semantic embedding learner.

\paragraph{Semantic alignment.} 
The second step is to build a visual semantic embedding model to align semantics in different modalities.
In this stage, $D_{S} = \{(x, y) \mid  x \in X_{S}, y \in  Y_{S} \}$ is used to train a semantic alignment learner.
This step can find structural or semantic correspondence between two embedding systems. 
It is the key technique to transfer semantic knowledge from word representations to a visual model.
In out framework, four independent paradigms (DeVISE, PrVISE, GrVISE and HyVISE) are defined below.
Each of them use different properties of word embedding systems. Note that the HyVISE which maps 
image and word representations to a Riemannian space has achieved the state-of-the-art performance in the standard ImageNet benchmark with traditional zero-shot setting.  
\begin{itemize}
  \item {\bf DeVISE}
  DeVISE is proposed by Frome \etal~\cite{frome2013devise} as the abbreviation of a deep visual semantic embedding model. In this work, we redefine it as a discriminative visual semantic model for its behaviour of 
  mapping visual features to the word embedding space by a combination of dot-product similarity and hinge rank loss.
  For a data instance in $D_{S}$, we have the loss:
  \begin{equation}
  \begin{split}
  &L_{DeVISE}(x, y) =   \sum_{j\neq y}\max[0, \\
  &margin - w(y)\cdot t(f(x)) +w_{j}\cdot t(f(x))]
  \end{split}
  \end{equation}
  where $t$ is a trainable transformation neural network. This loss aims to make image features close to their word embeddings and remains distances with word embeddings for incorrect labels.
  DeVISE is the natural way to project the information in the visual domain to the distributed word space.
  
  \item {\bf PrVISE} Probabilistic visual semantic embedding (PrVISE) is a model derived from a Variational Autoencoder~\cite{kingma2013auto} framework, a generative model with a prior distribution
   $p(z)=\mathcal{N} (z; 0, I)$. The VAE model makes it possible to approximate a latent data distribution from a reconstruction task. As its extension, PrVISE provides an adaptive latent prior distribution,
   which can be learnt by a neural network $q_{\phi_{w}}$ from the pre-trained word embedding space. By minimizing the KL divergence between the image latent distribution $q_{\phi_{i}}$ and the adaptive prior distribution, we define the loss of PrVISE as:
   \begin{equation}
  \begin{split}
  L_{PrVISE}(x, y) &= -\expect{ q_{\phi_{i}}(z|f(x)) }{ \logg{p_{\theta_{i}}(f(x)|z)}} \\
  &-\expect{ q_{\phi_{w}}(z|w(y)) }{ \logg{p_{\theta_{w}}(w(y)|z)}} \\
    &+ \kl{ q_{\phi_{i}}(z|f(x)) }{q_{\phi_{w}}(z|w(y))},
  \end{split}
  \end{equation} 
  where $p_{\theta_{w}}$ and $p_{\theta_{i}}$ are decoders for word and image feature reconstruction.
  PrVISE aims to find a joint embedding space for visual and linguistic information.
  
  \item {\bf GrVISE}
  GrVISE represents the visual semantic embedding model with Graph neural networks (like SGCN, DGP in~\cite{wang2018zero, kampffmeyer2019rethinking}).
  It addresses the problem that how to encode additional information in graph edges with a graph with weighted nodes. In the standard ZSL setting, it seems that capturing relational information from pre-trained word embeddings will augment semantics.
  In our case, GrVISE firstly trains a linear probe on $D_{S}$. 
  A shallow GCN $g(\cdot)$ is expected to distill discriminative visual semantic embeddings from the word embedding inputs. By minimizing the l2 loss between the GCN output $g(w)$ and the linear probe parameter $l$:
    \begin{equation}
  \begin{split}
  &L_{GrVISE} =   \sum_{w_{i}\in W, l_{i} \in L}{\left \| g(w_{i})-l_{i} \right \|^{2}}
  \end{split}
  \end{equation}
  we hope the GCN is able to predict the parameters of a new linear probe for unseen classes.
   
  \item {\bf HyVISE} HyVISE is a Riemannian version of DeVISE. Instead of mapping image features to a Euclidean space, HyVISE firstly projects image representations to a hyperbolic space with a map $e$, then models
  a Mobius transformer $m$ to make image representations close to their corresponding Poincar\'e embeddings. The loss is defined as:
  \begin{equation}
  \begin{split}
  &L_{HrVISE}(x, y) =   \sum_{j\neq y}\max[0,margin \\
  & + d(m(e(f(x))), p(y)) -d(m(e(f(x))), p_{j})]
  \end{split}
  \end{equation}
 \end{itemize}
Unlike previous works, we do not fine-tune the image encoder during the semantic alignment stage.
We consider this is essential to detect the effectiveness of each visual semantic embedding mechanism.

Closely related to our work is done by Sylvain~\etal~\cite{sylvain2019locality}, who propose a two-step
framework CM-DIM with self-supervised encoders. 
However, CM-DIM focused on the local and compositional behaviours of different pre-training representations for the conventional zero-shot learning benchmarks. 
Our intention is merely to explore how the role of distributed word representations are influenced in the different forms of zero-shot learning models, which we think is the key to make zero-shot learning more applicable in real-world applications. In this work, capturing image representations by a self-supervised method is not a prioritized problem, recent works~\cite{oord2018representation, tian2019contrastive, he2020momentum, chen2020simple} have demonstrated that contrastive learning can learn good image representations.

In the next section, we will use this unified ZSL framework and evaluate how a ZSL model works under a tough data split which needs high-level semantic information inference.

\section{Evaluation}
In this section, we firstly introduce the data split and evaluation tools and metrics. Then we try to answer the following 
questions with comparative experiments.
\begin{itemize}
    \item Does a visual semantic embedding model have the benefit to use information in the word embedding space?
    \item Do graph-based models learn the hierarchical distributed semantics?
\end{itemize}

\subsection{tieredImageNet}
The standard data split in the ImageNet has its own structural flaws. Test classes are the hypernym and hyponym  of training classes. On the other hand, testing images contain many inconsistencies, it is hard to ensure images with high quality are tested.
While tieredImageNet is a large subset of ILSVRC 2012 1k, in which classes are nodes of a tree of height
13 covering 608 classes. To build a ZSL benchmark, we combine the training classes and validation classes of tieredImageNet as the seen set, and we keep its own testing classes. 
In our ZSL setting, 448 classes are in the seen set, and 160 classes are in the testing set. Images in the training set of ILSVRC 2012 1k with the seen labels are used to train a visual semantic embedding model.
Images in the validation set of ILSVRC 2012 1k with the seen labels are used to evaluate embedding performance.
Images in the validation set of ILSVRC 2012 1k with the unseen labels are used to evaluate ZSL performance.
\subsection{Model Details}
For pre-trained image encoder, we use a ResNet-50 trained for 800 epochs on the ILSVRC 2012 1k with InfoMin CL Algorithm~\cite{tian2020makes} \footnote{https://github.com/HobbitLong/PyContrast}.
With a linear probe, this image encoder shows $73.0\%$ classification accuracy on the ILSVRC 2012 1k task.
We choose 300 dimensional GloVe\footnote{https://nlp.stanford.edu/projects/GloVe/} with 6B tokens as the pre-trained word vectors.
For each class label, we average all the GloVe vectors of its synonyms.
We train a Poincar\'e embedding model with gensim \footnote{https://radimrehurek.com/gensim/} from the tree of tieredImageNet classes. We choose a 100 dimensional hyperbolic space for this model.

The transformer in DeVISE is a two-layer multilayer perceptron (MLP) with 512 hidden neurons. The feature encoder and decoder in PrVISE are both two-layer MLPs with 512 hidden neurons.
The adaptive prior distribution in PrVISE is 300 dimensional, learnt by a two-layer MLP.
In GrVISE, a two-layer GCN is used to predict weights of the visual classifier.
And in HrVISE, a two-layer MLP is modified to implement Mobius multiplication twice.

All the visual semantic embedding models are trained with encoder freezing. The Adam optimizer~\cite{kingma2014adam} with learning rate 1e-4 is used in all the models to optimize the loss function.
All the models are trained for 200 epochs and mini-batches of size 256.

\begin{figure}[t]
\begin{center}
   \includegraphics[width=\columnwidth]{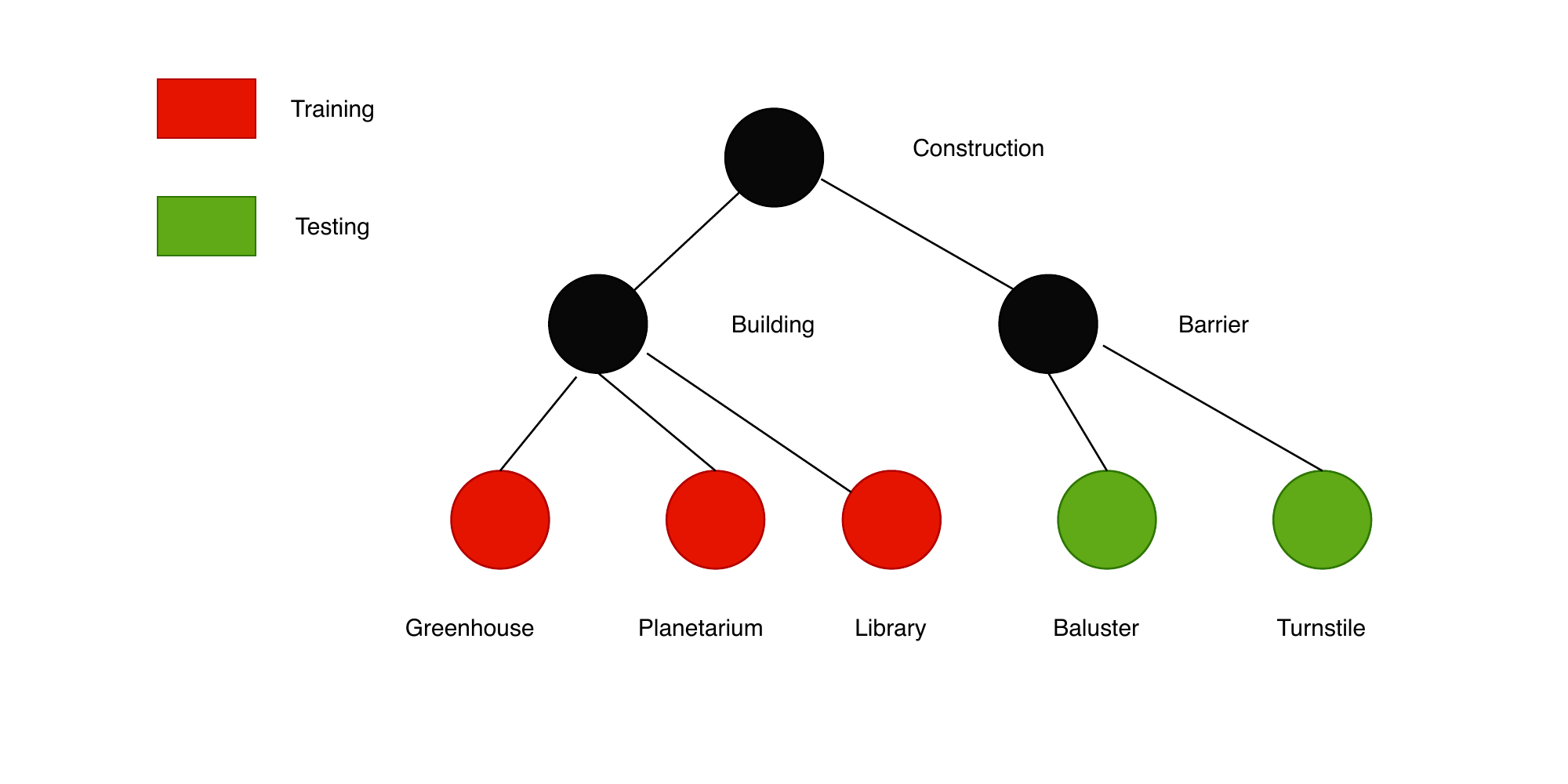}
\end{center}
   \caption{Illustration a subset of tieredImageNet. The training classes and test classes are nodes from different high-level semantic concepts. Compared with the standard split in Figure~\ref{fig:stand}, this spilt is more difficult for ZSL, while it provides complex semantic relationship and inference. We think it is an ideal way to demonstrate how a visual semantic embedding models works.}
\label{fig:tiered}
\end{figure}

\begin{table*}
\begin{center}
\begin{tabular}{l|l|l|l||l|l|l}
\hline
Models &  embedding hit@1 & embedding hit@5 & avg.sim@1 & avg.sim@5 & avg.sim.dis@1 & avg.sim.dis@5 \\
\hline\hline
LP & 75.18\%& 93.26\%& 0.8505& 0.5067& 21.80 & 69.13                \\
 
DeVISE & 71.96\%& 92.34\% & 0.8330& 0.5089 & 23.72 & 69.84            \\

PrVISE & 54.01\%& 74.24\% & 0.7353& 0.5087 & 29.25 & 52.51              \\
\hline
\end{tabular}
\end{center}
\caption{Comparison of model performance on visual semantic embedding task.
Only seen images in the testing set are used. A linear layer classifier LP performs better than the DeVISE and the PrVISE models on the flat hit metrics. The LP also achieve the highest semantic similarities for the wrong classified data. The DeVISE performs slightly worse than the LP. The VAE-based model PrVISE struggles to classify images by comparing the KL-divergence between latent distributions. This indicates a big performance gap exists when the latent space can not be disentangled.}
\label{table:1}
\end{table*}
\subsection{Metrics}
We consider different kinds of measures to answer the questions provided at the beginning of this section.

To demonstrate the performance of visual semantic embeddings, we compute the \textbf{embedding hit@k} metrics – the percentage of test images in the seen set for which the model returns the one true seen label in its top k predictions.
For the ZSL task, we consider the generalized ZSL. 
Therefore, we compute the \textbf{ZSL-S hit@k} \textbf{ZSL-U hit@k} metrics. Here S represents test images in  the seen set, while U represents  test images in  the unseen set.

Inspired by the average hierarchical distance of a mistake~\cite{bertinetto2020making}, which rethinks whether we should treat all classes other than the “true” label as equally wrong, 
we also employ the average similarity of a mistake \textbf{(avg.sim@k)} and the average similarity distance of a mistake \textbf{(avg.sim.dis@k)} as metrics to demonstrate whether a visual semantic embedding model remains the properties of a word embedding space.
To compute these two metrics, we firstly compute the cosine similarity matrix of the GloVe vectors.
Then we sort these similarities and get the distance of each pair of labels. 
For the data instances which the model does not return the true labels, we average the similarities and distances between the predicted labels and the true labels in their top $k$ predictions.
In the ZSL part, we also employ \textbf{ZSL-S avg.sim@k}, \textbf{ZSL-U avg.sim@k}, \textbf{ZSL-S avg.sim.dis@k} and \textbf{ZSL-U avg.sim.dis@k}.
The difference is we build the similarity matrices with the union set of training labels and testing labels.

We think the new metrics can help us figure out the effectiveness of semantic alignment in the ZSL task, as Roads and Love~\cite{roads2020learning} employed the Spearman
correlation between the upper diagonal portion of two similarity matrices in two conceptual systems.

\section{Experimental results}
\subsection{Does a visual semantic embedding model benefit from distributional semantics in a static word embedding space?}

To answer this question, we should consider the paradigms purely dealing with information from word embeddings.
In the following, we analyse the performance of DeVISE and PrVISE.
As a comparison, we implement a linear classifier (LP) on top of the frozen representation encoded by the InfoMin encoder.
Although the linear probe can not be used in the ZSL task, it still gives insights about how image representations are distributed with equally weighted labels.

The Table~\ref{table:1} shows results to evaluate visual semantic embedding performance.
We highlight that only seen classes in the testing set is used in this part.
A linear layer classifier performs better than the DeVISE and the PrVISE models on the flat hit metrics.
Although there are no distributed semantics leaked to the LP, it still shows the highest semantic similarities for the wrong
classified data.

DeVISE performs slightly worse than the linear classifier. There is no evidence to prove that 
pre-trained word embeddings help neural networks make a more reasonable decision at the semantic level, as random
embeddings still work well on the classification task~\cite{frome2013devise, bertinetto2020making}.

The generative model PrVISE shows the worst performance. 
Unlike the success on medium datasets with human designed attributes, the VAE-based model struggles to classify images 
by comparing the KL-divergence between latent distributions. This indicates a big performance gap exists when the 
latent space can not be disentangled.
However, the PrVISE model ``makes better semantic mistakes". 
Focusing on the difference of average similarities on the top 1 and top 5 predictions, 
these models do not prefer to choose candidate labels in the semantic synonyms of true labels. The PrVISE model does decrease the semantic distance of a mistake on the top 5 predictions.
Although, it is still very hard to answer the question ``Is visual similarity correlated to semantic similarity?”~\cite{deselaers2011visual}, it seems that the PrVISE model exchanges the embedding accuracy with the semantic accuracy. 
When we evaluate the performance of visual semantic embedding with extended labels, this phenomenon is more clear. DeVISE and PrVISE can deal with a flexible number of classes.
When the number of label classes increases, the embedding performance drops (see Table~\ref{table:2}), but the PrVISE model still ``makes better semantic mistakes". 
For unseen classes (Table~\ref{table:3}), PrVISE performs worse than DeVISE.
Notably, the performance in this work with contrastive learning pre-training reaches the same order of magnitude as the traditional setting, even though our data split is more difficult.

\begin{table}
\begin{center}
\begin{tabular}{l|l|l|l}
\hline
Models & LP& DeVISE& PrVISE \\
\hline \hline
ZSL-S hit@1&75.18\% &66.10\%& 53.97\%\\

ZSL-S hit@5 & 93.26\%&89.72\%& 74.16\%\\

ZSL-S avg.sim@1 & 0.8505&0.7857&0.7352\\

ZSL-S avg.sim@5 & 0.5067&0.4781&0.5091\\
ZSL-S avg.sim.dis@1 & 21.80 & 54.03& 37.93\\ 
ZSL-S avg.sim.dis@5 & 69.13&126.59& 67.89 \\
\hline
\end{tabular}
\end{center}
\caption{We test the visual semantic embedding models with full class semantics. This task maps seen image features into a space containing seen and unseen labels. Compared with Table~\ref{table:1}, DeVISE and PrVISE can deal with a flexible number of classes. When the number of label classes increases,  the embedding performance drops. The PrVISE model ``makes better semantic mistakes".}
\label{table:2}
\end{table}

\begin{table}
\begin{center}
\begin{tabular}{l|l|l|l}
\hline
Models & LP& DeVISE& PrVISE \\
\hline \hline
ZSL-U hit@1& N/A &1.590\%& 0.481\%\\

ZSL-U hit@5 & N/A&6.832\%& 2.806\%\\

ZSL-U avg.sim@1&N/A&0.3173&0.2805\\

ZSL-U avg.sim@5&N/A&0.3062&0.2760\\

ZSL-U avg.sim.dis@1&N/A&199.11 &198.30\\ 

ZSL-U avg.sim.dis@5&N/A&208.98& 204.18 \\
\hline
\end{tabular}
\end{center}
\caption{Comparison of the DeVISE and the PrVISE on the tiered-ImageNet ZSL split. Note that a linear layer can not predicts unseen classes. PrVISE performs worse than DeVISE. 
The results show that the performance with contrastive learning pre-training reaches the same order of magnitude as the traditional setting. Both DeVISE and PrVISE do not show the advantages, although the pre-trained features of unseen images are linearly separable.}
\label{table:3}
\end{table}

\subsection{Do graph-based models learn hierarchical distributed semantics?}

We test the performance of GrVISE and HyVISE to demonstrate how a semantic graph affects the way a model aligns visual semantics.
We discuss GrVISE and HyVISE respectively.

GrVISE is designed to predict the parameters of a linear image classifier with the help of GCN.
We set a two-layer MLP as a comparison.
In previous work using a GCN based model~\cite{wang2018zero}, parameters in the last layer of a pre-trained classifier are considered to be naturally normalized. However, we notice that the parameters in a pre-trained linear probe on a freezing encoder are not normalized. Therefore, we normalize the  parameters of the linear probe.
Note that the normalized parameters still work well the original classification task.  

\begin{table}
\begin{center}
\begin{tabular}{l|l|l}
\hline
Models & GCN& MLP\\
\hline \hline
embedding hit@1&29.67\% & 36.77\% \\

ZSL-S hit@1&29.63\% & 36.70\% \\

ZSL-U hit@1&0.832\% & 0.0\% \\
\hline
\end{tabular}
\end{center}
\caption{Results for the GrVISE on our ZSL setting. A two-layer MLP is also tested. The MLP achieves better parameter prediction for seen classes, while the graph-based model performs  better for zero-shot learning. As the same, they do not take the advantages of a semantic graph, when image features are linearly separable.}
\label{table:4}
\end{table}

We observe that an MLP performs better on parameter prediction.
Therefore, it achieves better embedding performance for seen classes. While a GCN performs better to predict unseen labels, this demonstrates that a GCN does learn a bit of structural information from a hand-crafted semantic graph.

However, the GrVISE does not organize visual semantic information better than the DeVISE.   
We argue that the process of learning a parameter prediction task can gradually  acquire semantic information for unseen classes. We train a linear probe on both seen and unseen classes. Then we train a GCN and an MLP respectively to predict the parameters of seen classes. Note that the probe can achieve $79.10\%$ classification accuracy on unseen classes.

\begin{figure}[t]
\begin{center}
   \includegraphics[width=\linewidth]{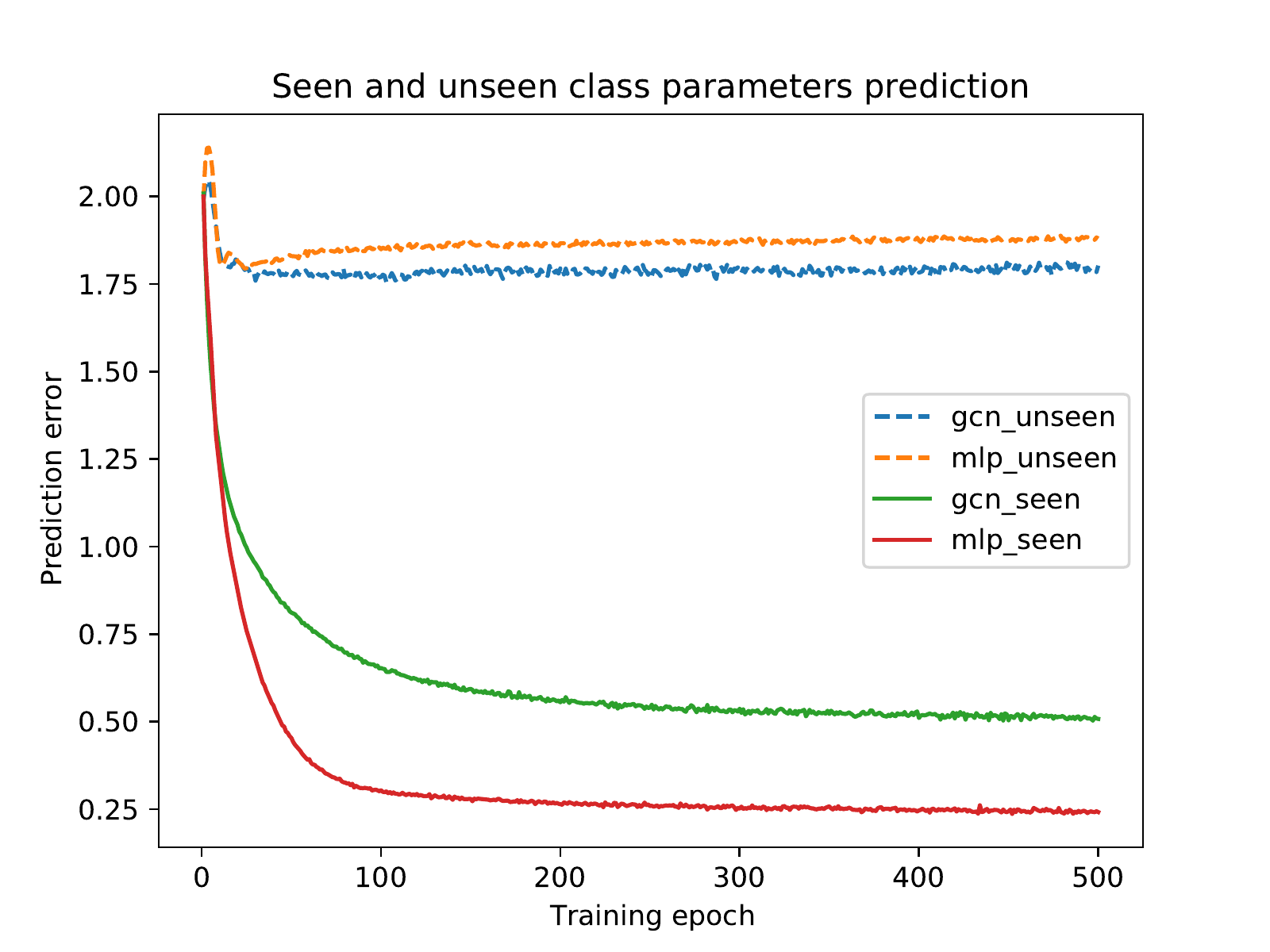}
\end{center}
   \caption{Illustration how a GCN and an MLP predict parameters of a pre-trained image classifier. Note that the dashed line belong to unseen classes. This demonstrates that the GCN achieves better parameter prediction on unseen image embedding.}
\label{fig:loss}
\end{figure}

In Figure~\ref{fig:loss}, we show that the GrVISE can not precisely learn unseen parameters.  The dashed lines representing the prediction error for unseen class parameters are observed stopping decreasing at the early epoch. The WordNet structure does not narrow the semantic gap between seen classes and unseen classes. 

Finally, we test the HyVISE, which can been seen as the DeVISE in Riemannian geometry.
We get 13.91\% embedding hit@1, 13.84\% ZSL-S hit@1 and 0.219\% ZSL-U hiy@1.
The results indicate that a Poincar\'e visual semantic embedding model does not work well on 
a complicate ZSL data split, when the split is based on a sub tree of the WordNet.

In summary, we find that the DeVISE still works robustly on a new ZSL class split, which needs high-level semantic understanding. 
Compared with a linear classifier, DeVISE does not show that it has performed semantic inference using the distributed semantics.
The generative model PrVISE performs slightly worse than the DeVISE, but it shows the advantage of predicting better mistakes, which are semantically similar to the ground truth labels. 
The graph-based models GrVISE and HyVISE lose their advantages shown the standard splits.
The results demonstrate that recent semantic alignment techniques could overfit to the original flawed problem, and they also indicate we need a deeper discussion on the ZSL task in the real world.

\section{Conclusion}
In this work, we investigate the problem of zero-shot learning.
We introduce the tieredImagenet split into ZSL to replace the standard data split.
That makes the evaluation of ZSL does not suffer from problems such as structural flaws, bad image quality.
We build a unified framework for ZSL with contrastive learning as pre-training. 
The two-step framework does not leak semantic information at the early stage, which makes us evaluate semantic inference fairly. 
With this framework, we test four mainstream methods for visual semantic alignment.
We demonstrate that current ZSL methods rely deeply on similarity comparison. 
No evidence shows that distributed visual information and distributed semantic information can be 
combined to implement semantic analogy and semantic inference across modalities.
Therefore, these ZSL methods can not generalize from complicated structural information. 

We believe that rethinking the goals and the definition of ZSL tasks is very much needed.
As Roads and Love~\cite{roads2020learning} pointed out in their work, it is unclear what mechanism makes
two-year-old children exhibit an average vocabulary of 200-300 words.
Learning a novel object is still the most challenging task.
During this decade, communities prefer to capture distributed knowledge from large image dataset and text corpus.
However, this method has been shown as an unrobust approach which can only works on hyponymy and hypernymy.
Context missing may be the main reason. 
Context information is more helpful than a distributed system.
Currently, vision-language pre-training has shown that it can effectively learn generic representations for downstream tasks.
In particular, Oscar~\cite{li2020oscar} has shown its success on captioning images with novel objects.
Context surrounding representations seem to outperform in capturing visual semantics~\cite{ilharco2020probing}.
We believe that it is necessary to build a ZSL task on contextual language models in the future to replace distributed word embedding models.
The new ZSL task will offer new practical insights on the way how human learn novel objects.
{\small
\bibliographystyle{ieee_fullname}
\bibliography{vse}
}

\end{document}